\documentclass[a4paper]{article}

\usepackage{amsmath}
\usepackage{graphicx}

\usepackage[style=numeric,backend=bibtex]{biblatex} 
\bibliography{image2vec_preblind}

\newcommand{\eg}{\emph{e.g.}, }       
\newcommand{\ie}{\emph{i.e.}, }      
\newcommand\etc{\emph{etc.}}

\title{A Visual Embedding for the Unsupervised Extraction of Abstract Semantics}
\author{Garcia-Gasulla, D. \& Ayguad\'{e}, E. \& Labarta, J.\\
Barcelona Supercomputing Center (BSC)\\dario.garcia@bsc.es\vspace{8pt}\\
B\'{e}jar, J. \& Cort\'{e}s, U.\\
Universitat Polit\`{e}cnica de Catalunya - BarcelonaTECH\vspace{8pt}\\
Suzumura T.\& Chen, R.\\
IBM T.J. Watson Research Center, USA}

\begin{document}
\maketitle

\begin{abstract}
Vector-space word representations obtained from neural network models have been shown to enable semantic operations based on vector arithmetic. In this paper, we explore the existence of similar information on vector representations of images. For that purpose we define a methodology to obtain large, sparse vector representations of image classes, and generate vectors through the state-of-the-art deep learning architecture GoogLeNet for 20K images obtained from ImageNet. We first evaluate the resultant vector-space semantics through its correlation with WordNet distances, and find vector distances to be strongly correlated with linguistic semantics. We then explore the location of images within the vector space, finding elements close in WordNet to be clustered together, regardless of significant visual variances (\eg 118 dog types). More surprisingly, we find that the space unsupervisedly separates complex classes without prior knowledge (\eg living things). Afterwards, we consider vector arithmetics. Although we are unable to obtain meaningful results on this regard, we discuss the various problem we encountered, and how we consider to solve them. Finally, we discuss the impact of our research for cognitive systems, focusing on the role of the architecture being used.
\end{abstract}

\section{Introduction}
Deep learning networks learn representations through the millions of features composing the network \cite{DLNature}. This provides a trained deep network with an exceptionally rich representation language, allowing it to perform detection and classification with remarkable precision. So far the representation language learnt by deep networks has been used straightforwardly, through tasks like image classification. However, deep network representations can be used for other purposes, if only the information coded within each feature can be extracted. A way of doing so is through vector-space representations. A vector-space that is then explored through vector arithmetics. Such is the approach taken by \cite{word2vec}, where authors find both syntactic (\eg singular/plural) and semantic (\eg male/female) regularities in vector representations of words.
In this paper we extract information from neural network models for the complex domain of images. This will lead us to work with deep networks capable of capturing the complexity and variety of information found on the visual domain. Using a previously trained network and its internal features as descriptors, we build vectors of features for a set of images using a trained network. Once the vector-space has been built, we analyze which semantics it contains. Results provide insight into the representations learnt by deep network models, and open up a new set of applications exploiting deep network representations.
\section{Motivation}
Word vector representations obtained from neural network models were found to contain syntactic and semantic information by \cite{word2vec}. This information can be extracted through arithmetic operations on the vector-space, and has been successfully used for tasks such as machine translation \cite{word2vecMT}. The motivation of this paper was to explore the existence of similar information in image vector representations, which could be useful for generic visual reasoning. 
Image vector representations extracted from convolutional neural networks (CNN) have been previously explored for their application to image recognition tasks. \cite{decaf} explored the performance of features learnt from a given data set at recognizing images classes from a different data set. These authors found that the top layer of a network seemed to cluster images according to high level semantics (\eg outdoor vs indoor). \cite{CNNFeatures} went further, considering the utility of these features to other image recognition problems such as fine-grained classification and attribute selection. Both of these works built image vector representations for solving image recognition tasks. Since the top layer of the network is optimized for discrimination during training, this layer turned out to be the most effective set of features available for the task. However, to represent abstract visual concepts (\eg image classes) which may or may not have been taught, the rest of the layers may become useful, as we try to maximize representativeness instead of discriminative power. Mid-level layers and parameters trained with one dataset were successfully reused for recognizing a different dataset by \cite{MIDTRANS}, showing the relevance of the learnt models.
A popular field of research right now within deep learning is multimodal systems, where visual and language models are integrated. An example of that is DeViSE \cite{DEVISE} which combines a skip-gram model trained on a large corpus and a CNN trained with ILSVRC data. Thanks to the information provided by the language model, DeViSE can make reasonable inferences on images belonging to unknown classes, in what is known as zero-shot prediction. A multimodel system particularly relevant for our work was proposed by \cite{MULTICAR}, combining an image-sentence embedding with a long short-term memory. Authors show the existence of regularities when performing operations such as \textit{image of a blue car - word blue + word red $\simeq$ image of a red car}. In this work we explore similar regularities, without using a language model to guide it. For that purpose we build sparse and high-dimensional representations of image classes, trying to obtain rich abstractions to empower this unsupervised process.
\section{Methodology}\label{sec:method}
\begin{figure}[!t]
\centering
\includegraphics[width=0.95\textwidth,natwidth=610,natheight=642]{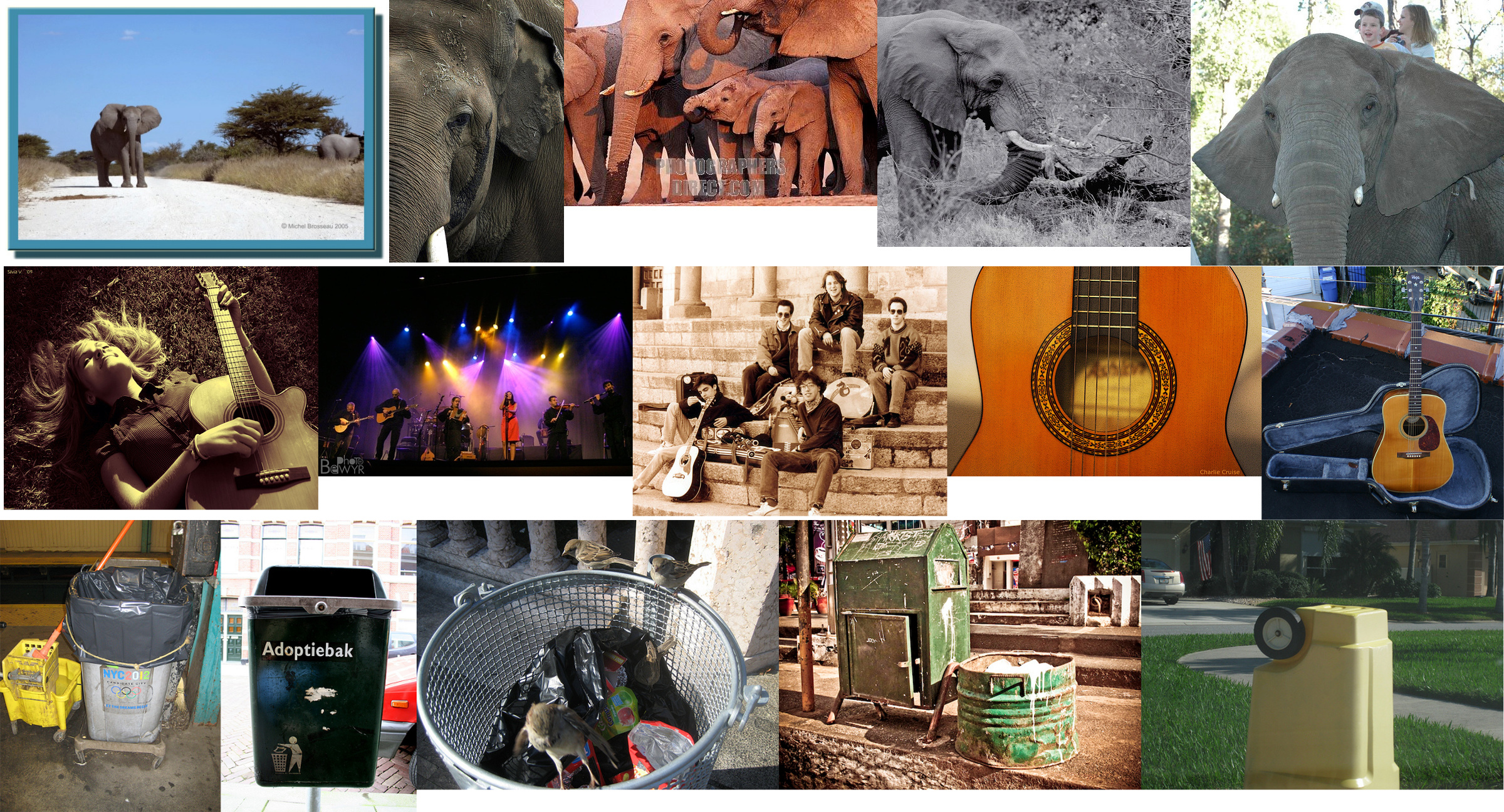}
   \caption{Sample of images used in the experiments, obtained from the ImageNet 2012 validation set. First row of images shown are labelled with the class \textit{"n02504458 African elephant, Loxodonta africana"}, the second row of images are labelled with the class \textit{"n02676566 acoustic guitar"}, and the third row of images are labelled as \textit{"n02747177 ashcan, trash can, garbage can, wastebin, ash bin, ash-bin, ashbin, dustbin,      trash barrel, trash bin"}.}
  \label{fig:sample}
\end{figure}

A CNN trained with labelled images learns visual patterns for discriminating those labels. In a deep network there can be millions of those patterns, implemented as activation functions (\eg ReLU) within the network features. Each feature within a deep network consequently provides a significant piece of visual information for the \emph{description} of images, even if they are not maximally relevant for their discrimination (only the top layer features are). By considering all feature activation values for a given image, one is in fact looking at everything the network \emph{sees} within the image, as \emph{learnt} from its training. Any visual semantics captured by the neural model will thus be found in those features values, values that we represent as a vector for their analysis.
The precision and specificity of a vector representation is bounded by the quality and variety of patterns found by the deep network; networks capable of discriminating more image classes with higher precision will provide richer image descriptions. To maximize both descriptive accuracy and detail we used the GoogLeNet architecture \cite{GOOGLENET}, a very deep CNN (22 layers) that won the ILSVRC14 visual recognition challenge \cite{ILSVRC15}. We used the pre-trained model available in the Caffe deep learning framework \cite{CAFFE}, trained with 1.2M images of the ImageNet test set for the task of discriminating the 1,000 ImageNet hierarchy categories.
The GoogLeNet model is composed by 9 \emph{Inception} modules. We capture the output of the 1x1, 3x3 and 5x5 convolution layers at the top of each of those 9 modules and build a vector representation with their activation values. When an image is run through the trained network, these 27 different layers combined produce over 1 Million activations, expressing the presence and relevance of as many different visual patterns in the input image. In our vector-building process we treat all composing features as independent variables. Thus, our image high-dimensional, sparse vector representation is composed by over 1M continuous variables. 
%
All executions described in this paper were performed on two Intel SandyBridge-EP E5-2670/1600 20M 8-core at 2.6 GHz and 64 GB of RAM. The code used to process the activation features and to produce all figures and graphs is available at https://github.com/dariogarcia/tiramisu. Our experiments use 20,000 images from the ImageNet validation set, which are labelled to 1,000 different classes, including a large variety of objects, animals, plants, etc. Each ImageNet class has a mapping to a different WordNet synset concept, which we will use to our advantage in the evaluation process (see \textit{Evaluation}). A sample of the images used is shown in Figure \ref{fig:sample}.

\subsection{Image Classes}
After obtaining vectors for 20,000 images, we perform an abstraction step to build vector representations of abstract classes, using the 1,000 classes images are labelled to. To build an image class vector we combine all the specific image vectors belonging to that class. As a result of this aggregation, we expect to obtain representative values of all variables for each class, reducing the variation found in specific images regarding brightness, context, scale \etc. The number of images aggregated per class ranges between 11 and 32. The aggregated image class vector has the same size as an image vector (roughly 1M variables), and is computed as the arithmetic mean of all images available for that class. At the end of this aggregation process we obtain 1,000 vectors, corresponding to the representations of each of the 1,000 leaf-node categories in the ImageNet hierarchy. Alternative aggregation methodologies were considered, as discussed in the \emph{Parametrization} section.
The 1,000 categories of the ImageNet hierarchy correspond to very diverse entities. Some of those are simple objects, producing few and weak activations. Others are more complex or found on rich contexts, involving more and stronger activations. This variability in the magnitude of image class vectors may affect the results of similarity metrics, since those classes having less and lower activations can be considered to be closer to other classes than what they actually are. A second source of variability is found in the variable behaviour of neurons given their location within the CNN. Typically, low-level neurons close to the input produce more frequent and stronger activations, as these represent simple patterns easier to find. On the other hand, neurons higher within the CNN produce more sparse activations, due to their specialized role. To eliminate the impact of both sources of variability we perform a normalization process on each image class vector. We normalize each vector values layer by layer, which guarantees that the information available at each visual resolution will be equally relevant for its representation, and that each image class vector will contain the same amount of information. Alternative normalization methods, including no normalization, were considered, as discussed in the \emph{Parametrization} section.
%
%
%
%
\begin{figure}[!ht]
\centering
\includegraphics[width=0.95\textwidth,natwidth=610,natheight=642]{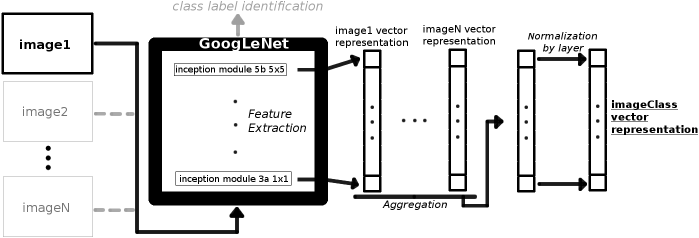}
   \caption{Feature extraction, aggregation and normalization process used to build image class vector representations.}
  \label{fig:process}
\end{figure}
%
%
%
Image class vector representations are built by aggregating and normalizing the activations of several images within a single vector, as depicted in Figure \ref{fig:process}. To study the information contained within the resultant vector-space we compute image class similarities through vector distance measures. We use the cosine similarity to build the distance matrix of the 1,000 classes, and use those distances for our vector-space evaluation (by comparing them with several distances based on WordNet) and analysis (by finding clusters of classes).



\section{Evaluation}\label{sec:eval}

To evaluate the consistency of the information captured by the proposed embedding space we use the labels of the represented classes. ImageNet labels are mapped to WordNet concepts, thus providing access to the \textit{lexical semantics} implemented in WordNet. Since vector representations are supposed to capture \textit{visual semantics} instead, a significant gap between both is to be expected. Nevertheless, WordNet remains the only source of validated knowledge available for evaluation.


Distances among image classes can be computed through WordNet measures, typically using the hypernym/hyponym lexical taxonomy \cite{wordnetSim}. At the same time we can compute image class distances in the vector-space, using the previously defined methodology. As a result we have, for every available image class, two sets of similarities with respect to the rest of image classes, similarities that can be reduced to a ranking. Spearman's $\rho$ provides a measure of correlation between two rankings, and is bounded between -1 and 1, with values close to either -1 or 1 indicating a strong correlation. We obtain a $\rho$ value for every image class, by comparing its lexical and visual rankings. By considering the $\rho$ values of all image classes we obtain a distribution of correlations, which indicates the level of semantic coherency between the WordNet taxonomy and the vector-space as a whole.

\begin{figure}[!t]
\centering
\includegraphics[width=0.95\textwidth,natwidth=610,natheight=642]{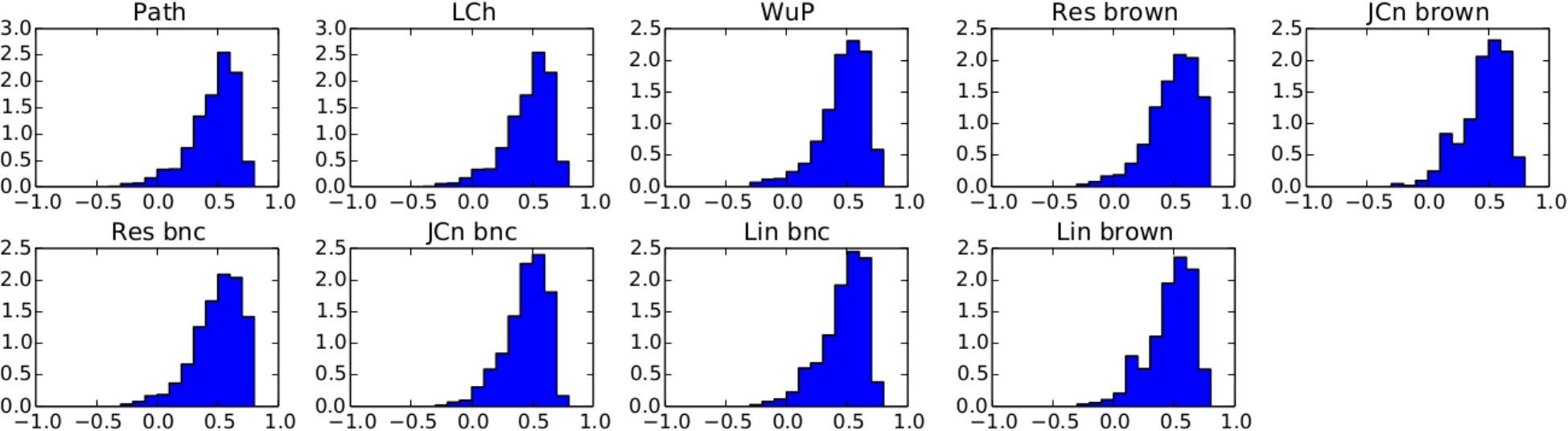}
   \caption{Histograms of Spearman's $\rho$ correlation between the image class vector similarity computed by our method, and nine WordNet similarity measures (three path length based: \emph{Path}, \emph{LCh} and \emph{WuP}, and three corpus-based \emph{Res}, \emph{JCn} and \emph{Lin} using two different corpus). Each histogram contains 1,000 $\rho$ correlation values corresponding to the correlation for every class in the ImageNet dataset.}
  \label{fig:hist}
\end{figure}

We consider six different WordNet distances to maximize consistency: three based on path length between concepts (\emph{Path}, \emph{LCh} and \emph{WuP}) and three corpus-based focused on the specificity of a concept (\emph{Res}, \emph{JCn} and \emph{Lin}) \cite{wordnetSim}. Additionally, we use two different corpus for the three corpus-based measures, the Brown Corpus, and the British National Corpus. Figure \ref{fig:hist} shows the distribution of correlations between the vector embedding and each of the nine WordNet measures. The $\rho$ values are mostly found between 0.4 and 0.6, indicating a strong correlation in a ranking with 999 elements. These results are consistent for all nine WordNet settings; the average $\rho$ on the distributions is 0.44 in the worse case (JCn bnc) and 0.49 in the best case (Res Brown \& Res bnc). This results indicate that vector representations contain a large amount of semantic information also captured by WordNet. This is a particularly interesting, considering that WordNet does not capture visual semantics such as color pattern, proportion or context.

\subsection{Parametrization}

Several alternative solutions were considered for the vector-building process described in the \emph{Methodology} section. To determine which specific solution was most appropriate we followed the same evaluation approach previously described. The options we considered were the following:

\begin{itemize}
 \item Aggregation: To compute the image class vector representation from a set of image vectors we used a mean. We considered the arithmetic, geometric and harmonic mean.

 \item Normalization: To normalize vectors we considered a normalization applied on the vector as a whole, and 27 sub-vector normalizations based on the 27 layers found on each vector. We considered the performance of both options applied to single image vectors, before aggregation, and to image class vectors, after aggregations. We also considered avoiding the normalization step.

 \item Distance: To compute distances between vectors we considered the cosine and euclidean distances. 

 \item Threshold: To decrease variability we considered adding an activation threshold to disregard activation values between 0 and 1 when building image vector representations, thus increasing vector sparsity.
\end{itemize}

We tested the combination of those parameters, evaluating them based on the distribution of correlations they produced. We found the best setting to be an arithmetic mean, an image class normalization by layer, cosine distance and no threshold. However, the imperfect nature of our evaluation (\ie using a ranking of correlations based in a lexical measure) does not allow us to definitely assert that this setting produces the most semantically rich vector embedding. Instead we discuss which parameters provided significant benefits in terms of correlation, and are thus recommended for future works, and which cause less definitive effects. The aggregation and normalization parameters had the largest impact on the distribution of correlations, allowing us to be confident on their setting. The arithmetic mean clearly outperformed the geometric and harmonic means, while the normalization both by layer and on the aggregated image class achieved much higher correlations than doing it either as a whole, in the original image vectors, or avoiding normalization all together. On the other hand, the distance algorithm, and particularly the use of an activation threshold, had a small impact on the distribution of correlations. For these parameters we choose the options which seemed to maximize correlations (cosine and no threshold), but different choices remain competitive.

A different parameter analyzed were the layers used to build the vector representation. Previous contributions argue it is best to consider only top layer activations when using them for image recognition tasks \cite{decaf,CNNFeatures}. Contrary to this approach, our methodology uses features from layers all over the network, with the goal of maximizing representativeness. To validate our notion, we use only certain layers of the network to build our embedding space, and then explore the correlations with WordNet on each setting. The distributions of correlations are not a fully comparable measure of quality, which is why we did not provide a formal evaluation of the previous parameters. However, since we need to provide some evidence contrasting with that of previous work \cite{decaf,CNNFeatures} regarding the layers being used, we decided to show the mean $\rho$ obtained by each subset of layers. The values in Table \ref{tab:rhos} are not to be taken as definitive evidence. 

We consider using only the features belonging to the top 22\% of the network (\ie inception modules 5a and 5b), features from the middle 55\% (\ie inception modules 4a, 4b, 4c, 4d and 4e) and features from the bottom 22\% of the network (\ie inception modules 3a and 3b). Results indicate that the correlation achieved by the middle 55\% is similar to the correlation achieved by the set of all 27 layers. On the other hand, the correlations achieved when considering only features from the top 22\% or the bottom 22\% layers were both significantly worse, while still showing correlation. These results indicate that all layers within the network contain visual semantics relevant for the description of abstract image classes, regardless of their location.

\begin{table}[t]
 \centering
  \begin{tabular}{|l|c|c|c|c|}
    \hline
    \textbf{Layers used}	& All layers	&Top 22\%	&Middle 55\%	&Bottom 22\%	\\ \hline
    \textbf{Mean $\rho$}	&0.46		&0.41		&0.46		&0.36\\ \hline
  \end{tabular}
\caption{Mean $\rho$ correlation values between the image class vector similarity computed by our method, and nine WordNet similarity measures, when using a subset of layers of the CNN model.}
\label{tab:rhos}
\end{table}

The differences between these results and those of previous works \cite{decaf,CNNFeatures} are explained by the differences in our methods and goals. Previous contributions focused on single image representations for image recognition tasks. Since single images are highly variable in terms of brightness, context, \etc, the use of the more disperse lower layers for their representation maybe counterproductive. We on the other hand target high-level image class representations, which have a much smaller variability thanks to the aggregation and normalization processes we apply. As a result we can consider larger and more volatile parts of the input (\ie the non-top layers) which seem to be potentially useful for the knowledge representation process when targeting abstract entities.


\section{Clusters of Image Classes}

\begin{figure}[!t]
\includegraphics[width=0.95\textwidth,natwidth=610,natheight=642]{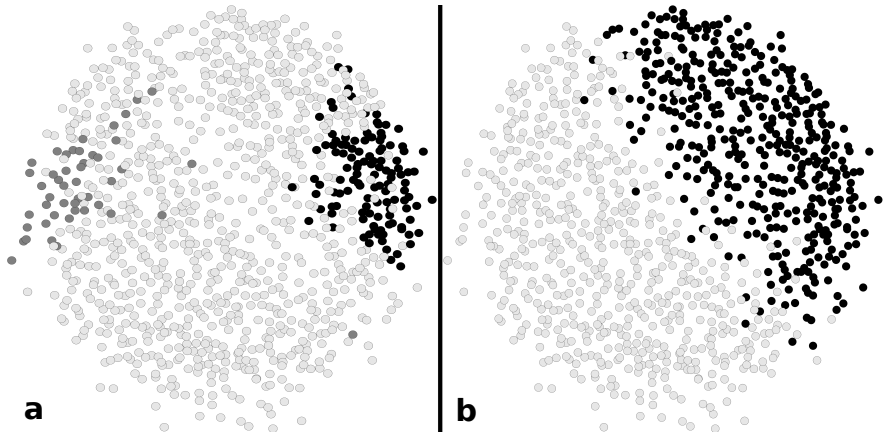}
   \caption{Scatter plot of image class vector similarities built through metric multi-dimensional scaling. (a) black circles belong to images labelled as hyponyms of synset \textit{"dog, domestic dog, Canis familiaris"} and dark grey circles belong to images labelled as hyponyms of synset \textit{"wheeled vehicle"}. (b) black circles belong to images labelled as hyponyms of synset \emph{"living thing, animate thing"}.}
  \label{fig:graph_both}
\end{figure}

To further analyze the semantics captured within the defined vector-space we perform a supervised analysis of clusters, using the WordNet hierarchy as ground truth; by knowing which image classes are hyponyms of the same synset, we can explore their distribution within the embedded space. To achieve visual results, we apply metric multi-dimensional scaling \cite{Borg2005} with two dimensions on the 1,000 image classes distance matrix. This method builds a two-dimensional mapping of the vector distances which respects the pairwise original similarities. We first use two synsets with many hyponyms within the ImageNet categories: \emph{dog} (according to WordNet there are 118 specializations of \emph{dog} in the image classes) and \emph{wheeled vehicle} (with 44 specializations of \emph{wheeled vehicle} in the image classes\footnote{To these 44 classes we added the \emph{school bus}, \emph{minibus} and \emph{trolleybus} image classes, which we consider to be wheeled vehicles.}). We highlight the location of the image classes belonging to each one of these two sets in the two-dimensional similarity mapping of Figure \ref{fig:graph_both}a. 

At first sight, the two sets of highlighted images compose definable clusters. Although precision is not perfect, image classes belonging to the same WordNet category are clearly assembled together in the vector-space representation. In the case of dogs, this is relevant because of the wide variety of dogs computed, some of which have few visual features in common (\eg \emph{Chihuahua}, \emph{Husky}, \emph{Poodle}, \emph{Great Dane}). According to these results, the visual features which are common on all dogs have more weight on the vector representation than variable features such as size, color or proportion. This is probably caused by the aggregation and normalization process, which reduces the importance within image classes of volatile properties. The cluster defined by wheeled vehicle image classes has a lower precision than that of dogs, probably because wheeled vehicles are more varied than dogs (\eg \emph{Monocycle}, \emph{Tank}, \emph{Train}). Nevertheless all but one wheeled vehicle are located on the same quadrant of the graph, indicating that there is a large and reliable set of features in the vector representation identifying this type of image classes. The one wheeled vehicle located outside of the middle-left quadrant, in the low-right part of Figure \ref{fig:graph_both}a, corresponds to \emph{snowmobile}, a rather special type of wheeled vehicle which seems to be different to everything.

\begin{figure}[!t]
    \centering
\includegraphics[width=0.95\textwidth,natwidth=610,natheight=642]{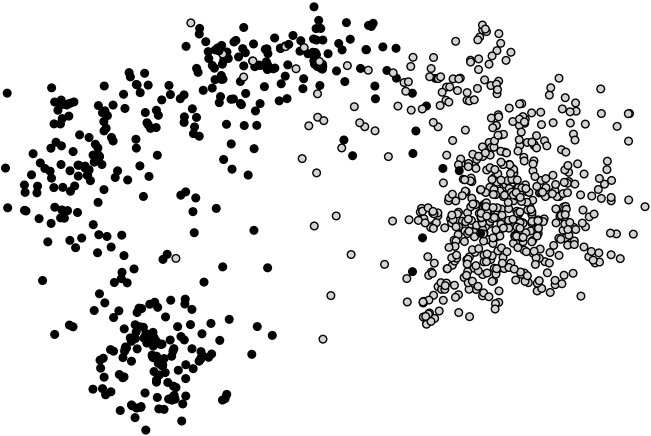}
    \caption{Scatter plot built through ISOMAP, analogous to Figure \ref{fig:graph_both}b. Black circles belong to images labelled as hyponyms of synset \emph{"living thing, animate thing"}.}
    \label{fig:graph_living_things_ISO}
\end{figure}

By looking at Figure \ref{fig:graph_both}a we notice a gap naturally splitting image classes into two sets. This separation is the only consistently sparse area visible at first sight in the graph. To explain this phenomenon we explored the most basic categorization in WordNet, separating ImageNet classes between living things, defined by WordNet as \emph{a living (or once living) entity} and the rest. By painting the images belonging to living things we obtain the graph of Figure \ref{fig:graph_both}b. This graph shows how the separation found in the vector-space corresponds to this simple categorization with striking precision, unsupervisedly clustering images depending on whether they depict living things or not. The few mistakes done correspond to organisms with unique shapes and textures (\eg lobster, baseball player, dragonfly) and things which are often depicted around living things (\eg snorkel, dog sled). Other particular cases are rather controversial, as \emph{coral reef} is not a living organism according to WordNet but in the vector-space it is clustered as such. Encouraged by these results we tried to obtain a representation of the vector-space which showed the separation between living organisms and the rest with more clarity. For that purpose we tested a non-linear mapping of the distances to three dimensions using the ISOMAP algorithm \cite{Tenenbaum2000}. The features extracted from the network are combined non-linearly to obtain the class of an image, so this kind of transformation should highlight the inherent non-linearity of the vector-space. Figure \ref{fig:graph_living_things_ISO} shows a more evident separation among these two synsets and also a more complex structure within the class images.

At this point we can assert that vector representations capture large amounts of high-level semantics. Given that the source deep network was only provided with 1,000 independent image category labels, all the semantics captured beyond those in the vector space must originate from visual features. The boundaries of what semantics can be captured this way are however hard to define, as it would require us to state what can and cannot be learnt only from seen images. Our best notion so far on the limits of visual semantics is provided by the distinction between living things and the rest. Horses, salmons, eagles, lizards and mushrooms seem to have little in common visually, and yet these elements are clustered in the vector-space. The structural patterns of living things seem therefore to be particular enough as to motivate a distinction. These results open up many interesting questions which we intend to address as follow-up work.

\section{Lessons on Image Equations}
Word vector embeddings empower the extraction of syntactic and semantic regularities through vector arithmetics (\eg ``King - Man + Woman = Queen'') \cite{word2vec}. Most of the regularities explored in the linguistic context so far have been of the form ``$a$ is to $a^*$ as $b$ is to $b^*$'', having applications in fields like machine translation \cite{word2vecMT}.

The image class vector embedding construction process defined in the \textit{Methodology} section allows us to consider the existence of similar regularities within the domain of images. If, as the results from the \textit{Evaluation} section indicate, visual semantics are encoded within our image class vectors, it may be possible to operate with those semantics to perform some type of visual arithmetic reasoning. In comparison with the linguistic context, the image embedding space we build has a much larger dimensionality and data variability. To simplify the arithmetic process we consider simpler relations with only three operands, of the form ``$a$ - $b$ $\simeq$ $c$''. We use the subtraction operator on image class vectors, combining it with the cosine similarity measure as the $\simeq$ operator to find the closest image class to the result of the substraction and thus solving the equation. Given two images, $i_1$ and $i_2$ and a feature $f$, the subtraction of $i_2$ from $i_1$ for $f$ is defined as

\[
f(i_1 - i_2) =
\begin{cases}
f(i_1) - f(i_2), & \text{if $f(i_1) > f(i_2)$}\\
0, & \text{otherwise} 
\end{cases}
\]

The substraction operation is defined here at a feature level. To solve image equations we substract at vector level by applying it of all the features composing the vector.

We consider two different scenarios where these equations could be applied. One is to analyze image classes which can be understood as the non-overlapped concatenation of two other classes. An example of that could be \emph{chair} plus \emph{wheel}, which could produce image classes like \emph{office chair} or \emph{wheelchair}. The second scenario we foresee is to analyze image classes which can be understood as the overlapped combination of two other classes. In this case the visual properties of the two mixed classes would need to be strongly intertwined to produce a third. An example of that could be \emph{wolf} plus \emph{man}, which could produce \emph{werewolf}.

When exploring the image equations here described we found several limitations which prevented us from obtaining meaningful results. Since this is a work in process we describe these limitations here, and how we intend to solve them, hoping that other researchers may benefit from our experience. The first problem we faced was related to the evaluation of image equations, since there is no obvious ground truth available. One could subjectively propose many different equations which make visual sense, such as \textit{platypus - duck = beaver}, \textit{turtle - shield = lizard} or \textit{motorcycle - motor = bicycle}. This approach, besides being hardly scalable since one's imagination quickly runs out, may easily include a human bias through the implicit consideration of non-visual features. Nevertheless, this remains the best available evaluation methodology.

Since there is no available test dataset, we can instead compute all possible equations, and subjectively decide if the most exact ones (those with a lower $\simeq$ operator value) make sense visually or not. The problem with this approach is the huge number of possible equations there are, and the computational cost of computing them all. For example, using the 1,000 different class vectors we obtained from the ImageNet dataset, we can compose roughly 997 million different equations of the form ``$a$ - $b$ $\simeq$ $c$'' (notice the cosine similarity is commutative and the substraction operator is not). To compute a single equation one needs to substract two vectors composed by a one million features, and then compute the similarity of the result with a third vector. Computing 997 million of these is therefore a problem requiring huge amount of computational resources. Thus, very efficient code running in parallel on top of high-performance infrastructure is a must for solving this problem.

Another relevant problem is caused by the curse of dimensionality. The image class vector representations we work with are composed by roughly one million features. Consequently, the resultant vector space is defined by one million dimensions. In such a high-dimensional space all objects will be highly similar, and distance measures, such as the euclidean distance, will have trouble finding meaningful differences. To solve this issue we are considering various dimensionality reduction methods (e.g., principal components analysis), as well as employing distance measures which are more resistant to high dimensionality (see \cite{aggarwal2001surprising}).

\section{Implications for Cognitive Systems}

Deep learning networks have been shown to be most appropriate to solve a specific component of cognitive systems: perception. Unlike previous models, deep networks can process huge amounts of raw data, first learning and later identifying abstract patterns characterizing the data for complex tasks. Through its multiple non-linearities, deep learning models solve the cognitive problem of transforming sub-symbolic data into symbolic knowledge; the difference between looking and seeing. In the case of CNNs, these models mimic the animal visual cortex by processing images as a two dimensional matrix of features. Squared sub-parts of the matrix are processed by receptive fields, which look for previously learnt patterns or filters all over the input picture (\ie a convolution). The output of each filter is projected into two dimensional matrix, with the goal of iteratively applying the same convolution process to a larger part of the original input matrix. Eventually, neurons are fed with the full input matrix, producing data descriptors of the whole input. Current deep CNN architectures include millions of filters distributed among several layers, each of these filters tuned for a given purpose. This design makes CNNs see as humans do in essence, perceiving small visual patterns on small visual patches of the input, and iteratively aggregating these until full visual perception is obtained.

Although deep learning apparently solves the problem of perception through the definition of a large and rich representation language, these models seem to be limited for higher-level purposes related with cognition (\eg reasoning) as they lack symbol operating mechanisms. This situation naturally leads to a solution where additional machine learning methods can be plugged on top of a deep learning systems, processing the representations built by these to achieve high-level cognitive processes. This particular architecture is the one we follow with our research, where we use the representation language learnt by a CNN (a pretrained GoogLeNet model) to obtain a specification of the input data (image vectors and class vectors) to empower additional machine learning methods (clustering algorithms). Using this architecture we are capable of performing high-level cognition such as \textit{concept discovery}, something CNNs on their own cannot do, by finding clusters of entities which correspond to entities (\eg \textit{living thing}). 

It is currently impossible to say which may be the overall impact of deep learning for cognitive systems. We can however assert that at least the lowest level of cognition, perception, can be replicated through models such as CNNs. The same combined architecture we propose is in essence behind most deep learning success cases (\eg deep learning + classifier for object recognition \cite{he2015delving}, deep learning + tree search and reinforcement learning for boardgames and video games \cite{silver2016mastering,mnih2013playing}), and it is most likely that the next generation of cognitive systems will follow this architecture as well.

\section{Conclusions}

In this paper we present a methodology to build vector representations of image classes based on features originally learnt by a deep learning network. Our goal was to extract the visual semantics captured by the deep network model, in order to make them available to other learning and reasoning methods. Unlike previous research, we focus on representing abstract classes, by aggregating and normalizing single images belonging to the same concept. The consistency of the methodology allows us to consider an unprecedented amount of features (over 1M). 

We analyze the resultant vector-space first by looking at the clustering of different elements with common semantics (\eg dogs, wheeled vehicles). Variations in proportion, size and color within those classes seems to be dominated by more essential visual features (\eg those shared by 118 kinds of dogs), as these elements are clustered together within the embedding space. The existence of high-level properties is further supported by an untaught vector distinction between living organisms and non-living things. This makes us wonder what is the limit of what can be learnt through visual information, as it is not straightforward to define which are the visual particularities of life. 


In general terms, the proposed methodology takes a large volume of pixels (sub-symbolic data) and translates them into concepts with abstract semantics (symbolic knowledge). This workflow has clear implications for artificial cognitive systems, as it tackles the problem of obtaining symbolic knowledge from sub-symbolic data, a necessary step to abstract reasoning through the real world perception. In practice, the approach extracts high-level knowledge of images through deep networks, making the representational power a CNN available for other cognitive purposes. A promising architecture for cognitive systems. Identifying clusters of images distinct in the vector-space, or finding the distinctive traits of a set of classes could be used for visual learning, while vector arithmetics could be used for reasoning and artificial image generation. The main follow-up work of this research goes in that direction, solving the problems we found when applying vector arithmetics, and exploring how to use deep learning representations outside of deep learning. We also intend to apply our work to the zero-shot prediction task, as this approach would allow us to tackle it in a completely unsupervised manner.

\section{Acknowledgements}
This work is partially supported by the Joint Study Agreement no. W156463 under the IBM/BSC Deep Learning Center agreement, by the Spanish Government through Programa Severo Ochoa (SEV-2015-0493), by the Spanish Ministry of Science and Technology through TIN2015-65316-P project and by the Generalitat de Catalunya (contracts 2014-SGR-1051), and by the Core Research for Evolutional Science and Technology (CREST) program of Japan Science and Technology Agency (JST).

\printbibliography    

\end{document}